\documentclass{article}
\usepackage{spconf,amsmath,graphicx}

\usepackage{epsfig}
\usepackage{amssymb}
\usepackage[linesnumbered,vlined,ruled]{algorithm2e}

\title{UNSUPERVISED MEDICAL IMAGE ALIGNMENT WITH CURRICULUM LEARNING}
\name{Mihail Burduja \qquad {Radu Tudor} Ionescu}%\thanks{Thanks to XYZ agency for funding.}}
\address{Faculty of Mathematics and Computer Science, University of Bucharest, Romania}
\begin{document}
%\ninept
%
\maketitle
% Abstract must be up to 150 words
\begin{abstract}
We explore different curriculum learning methods for training convolutional neural networks on the task of deformable pairwise 3D medical image registration. 
To the best of our knowledge, we are the first to attempt to improve performance by training medical image registration models using curriculum learning, starting from an easy training setup in the first training stages, and gradually increasing the complexity of the setup. On the one hand, we consider two existing curriculum learning approaches, namely curriculum dropout and curriculum by smoothing. On the other hand, we propose a novel and simple strategy to achieve curriculum, namely to use purposely blurred images at the beginning, then gradually transit to sharper images in the later training stages. Our experiments with an underlying state-of-the-art deep learning model show that curriculum learning can lead to superior results compared to conventional training. Additionally, we show that curriculum by input blur has the best accuracy versus speed trade-off among the compared curriculum learning approaches.
\end{abstract}

\setlength{\abovedisplayskip}{4.0pt}
\setlength{\belowdisplayskip}{4.0pt}

\section{Introduction}
\label{sec:intro}
\vspace{-0.15cm}

\emph{Image registration} is a smooth alignment between two images that maps point coordinates from one image to corresponding point coordinates in the second image. \emph{Medical image registration} maps point coordinates of some anatomical structures from the first image onto point coordinates from the same anatomical structures in the second image. Medical image registration is one of the most studied problems in medical image analysis \cite{ashburner2007fast,balakrishnan2018unsupervised,brown1992survey,hill2001medical, oliveira2014medical,zhao2019unsupervised, zhao2019recursive}, which helps with the alignment and fusion of scans taken through different means, e.g.~computer tomography (CT) or magnetic resonance imaging (MRI), or at different times, leading to better computer assisted diagnosis among other benefits.

The state-of-the-art methods for medical image registration are based on deep neural networks~\cite{balakrishnan2018unsupervised,zhao2019unsupervised, zhao2019recursive}. These models are usually trained using a variant of stochastic gradient descent on mini-batches that are randomly selected from the training set. However, the conventional training based on random data selection is not always optimal. As noted by Bengio et al.~\cite{bengio2009curriculum}, curriculum learning represents a training strategy from easy to hard, which may guide neural models to better local optima. To this end, we investigate a series of curriculum learning strategies for unsupervised medical image registration. Besides considering state-of-the-art curriculum learning approaches such as curriculum dropout \cite{morerio2017curriculum} and curriculum by smoothing \cite{sinha2020curriculum} for our experiments, we propose a novel curriculum learning strategy that starts the training process on intentionally blurred images and then makes a gradual transition to sharper images in the later training stages.

We perform our evaluation on the SLIVER \cite{heimann2009comparison} data set in order to compare the proposed curriculum learning methods with conventional training. The underlying model in our experiments is the state-of-the-art recursive cascade network proposed by Zhao et al.~\cite{zhao2019recursive}. The empirical results indicate that curriculum learning can lead to superior performance compared to the conventional learning process based on randomly chosen mini-batches.

% The paper is organized as follows. Related work is discussed in Section~\ref{sec:relatedart}. The curriculum learning approaches are presented in Section~\ref{sec:method}. The experiments are discussed in Section~\ref{sec:experiments}. We draw our conclusions in Section~\ref{sec:conclusion}.

\vspace{-0.25cm}
\section{Related work}
\label{sec:relatedart}
\vspace{-0.15cm}

Image registration has been studied extensively in literature \cite{ashburner2007fast, balakrishnan2018unsupervised,brown1992survey,hill2001medical, oliveira2014medical, zhao2019unsupervised, zhao2019recursive}. Medical image registration methods are used for the alignment and fusion of different types of scans (CT, PET, MRI, etc.) of the same anatomical structure, being useful in computer aided diagnosis, computer assisted surgery and treatment. Recent approaches use convolutional neural networks (CNNs) trained specifically for registration \cite{balakrishnan2018unsupervised,zhao2019unsupervised, zhao2019recursive,jaderberg2015spatial}. One of the first networks that was used for image registration is a spatial transformer network (STN)~\cite{jaderberg2015spatial}, which was not originally proposed as a method for image registration, requiring a certain degree of adaptation.

\noindent
{\bf Engineered image registration methods.}
Early methods for image registration were based on solving the differential equations that morph one image into another~\cite{christensen19943d}.  DARTEL~\cite{ashburner2007fast} is a method that uses a single flow field and is computationally more efficient than methods that use multiple flow fields. Moreover, the resulting deformations are easily invertible. The iterative closest point algorithm~\cite{besl1992method} represents another method for 3D image registration. The method employs the iterative closest point algorithm to handle the full six degrees of freedom. Other early approaches are intensity-based or feature-based methods, usually relying on handcrafted features.
The main drawback of these methods~\cite{ashburner2007fast, beg2005computing, vaillant2004statistics} is that the involved differential equations are difficult to work with, leading to suboptimal results in realistic scenarios that entail non-rigid distortions.

\noindent
{\bf Trainable image registration methods.} The lack of reliance on handcrafted features and the vast ability of deep neural networks to learn patterns of anatomical structures makes them good candidates for the task of medical image registration. The research on image registration based on deep neural networks is rich~\cite{dosovitskiy2015flownet, krebs2017robust, rohe2017svf}, and it usually involves supervised models that learn from ground-truth labels provided in the form of segmentations or warp fields. However, medical data containing ground-truth labels is scarce, mainly due to the laborious annotation process that is required. Hence, the potential of using supervised learning for medical image registration is limited. To this end, several recent works proposed unsupervised frameworks that use data sets~\cite{dosovitskiy2015flownet, mayer2016large} with synthetic annotations and similarity-based loss functions~\cite{balakrishnan2018unsupervised,zhao2019unsupervised, zhao2019recursive}. Similar to these studies, we consider an unsupervised framework. Unlike prior works on medical image registration, we study curriculum learning approaches to improve the results of unsupervised models for medical image registration. 

\noindent
{\bf Curriculum learning.} Curriculum learning was introduced by Bengio et al.~\cite{bengio2009curriculum} as a technique to optimize machine learning models. It is based on the fact that neural networks are inspired by the human brains and they should be trained the same way humans learn -- starting with easy examples and progressively increasing their complexity. Since its introduction, curriculum learning was applied on a wide range of deep learning problems, as noted in the recent survey of Soviany et al.~\cite{soviany2021curriculum}. %There are also training methods that oppose curriculum learning, starting the training process with the more difficult examples~\cite{shrivastava2016training}.
A problem of the original (data-level) curriculum learning formulation is that the training examples must be ranked according to their complexity~\cite{soviany2021curriculum}, preventing the application on tasks and data sets for which the complexity of the examples is unknown or cannot be determined. While this situation is no longer problematic for natural images since the development of image difficulty estimators~\cite{Ionescu-CVPR-2016}, we are still confronted with the problem when it comes to medical images.
%while for some tasks ranking according to complexity may be accessible, data sets with large amounts of data and complex structure are difficult to rank. 
To avoid this problem, some (model-level) curriculum learning methods turned to alternative strategies such as gradually increasing the model's complexity~\cite{karras2017progressive}, decreasing the dropout rate~\cite{morerio2017curriculum} or smoothing the convolutional filters~\cite{sinha2020curriculum}. %For example, PROGAN~\cite{karras2017progressive} uses different image resolutions at different stages of training, starting with $4\times4$ images and gradually increasing the resolution and the capacity of the model. 
These methods provide promising results when the difficulty of the data samples is hard to determine. In this work, we propose a curriculum learning strategy that artificially simplifies the examples in the initial training stages through blurring. Our method can be regarded as a data-level curriculum technique which, unlike other data-level strategies, does not require a difficulty estimator for the data samples.

\vspace{-0.25cm}
\section{Method}
\label{sec:method}
\vspace{-0.15cm}

The \emph{Volume Tweening Network} (VTN) was introduced by Zhao et al.~\cite{zhao2019unsupervised} as an unsupervised end-to-end framework that uses CNNs for 3D medical image registration. The VTN is based on a set of stacked registration sub-networks, achieving state-of-the-art results by predicting a dense flow field using deconvolutional layers. There are two types of sub-networks used in the VTN, namely affine networks and dense deformable registration networks. The affine network aims to perform an initial alignment, and is only used once, as the first sub-network. The dense deformable networks are based on an encoder-decoder architecture that receives as input the fixed image and the current moving image (obtained from previous sub-networks in the stack). The encoder part is composed of convolutional layers and the decoder is composed of deconvolutional layers, similar to the U-Net architecture~\cite{ronneberger2015u}. The output of each sub-network is a dense flow field that contains three-axis displacements of the same size as the input. As the underlying model for our study, we employ the 1-cascade VTN network~\cite{zhao2019unsupervised, zhao2019recursive} and augment it as needed in order to apply a curriculum learning strategy or the other.

\begin{figure}[!t]
\begin{center}
\includegraphics[width=1.0\linewidth]{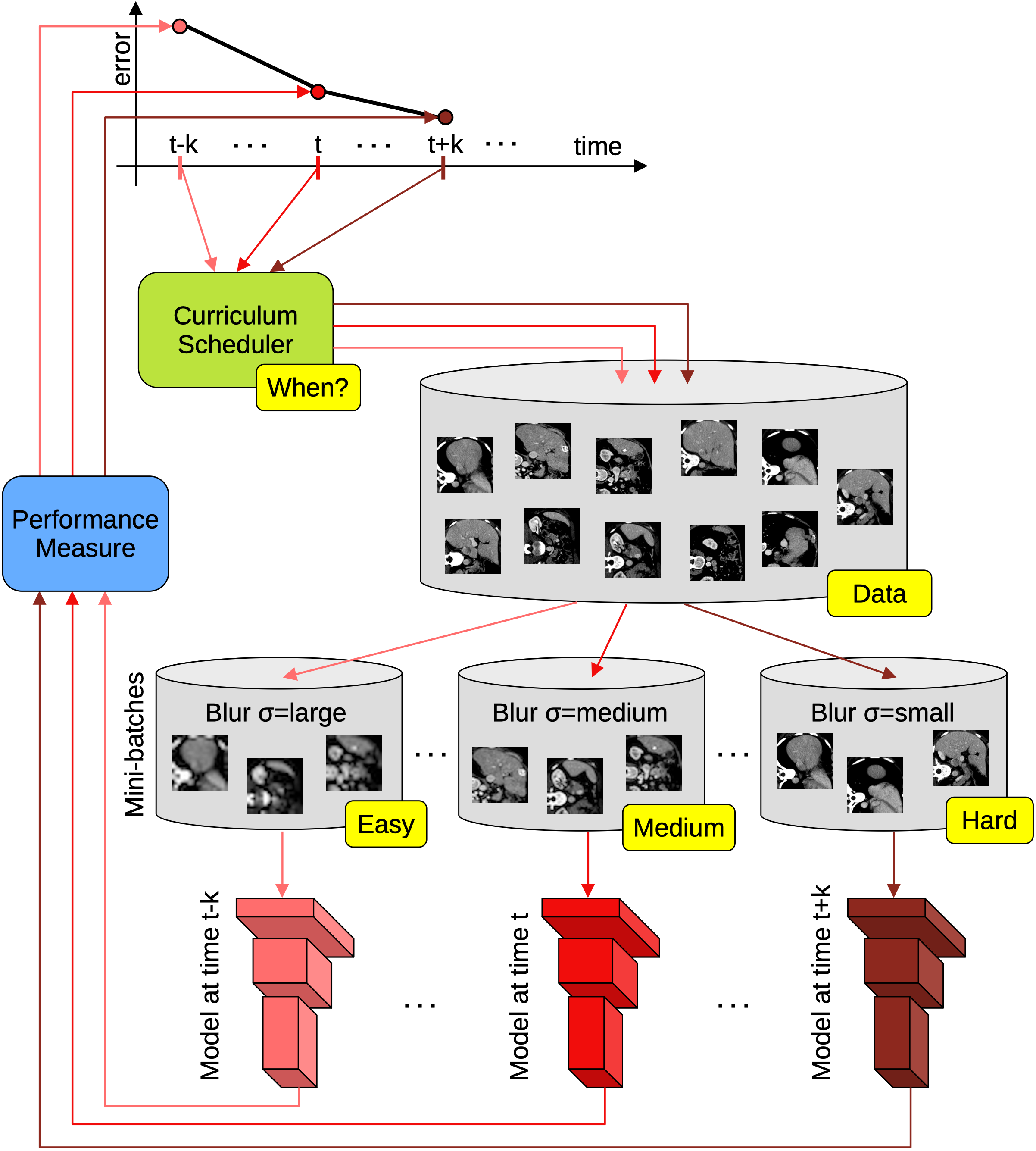}
\end{center}
\vspace*{-0.7cm}
\caption{The proposed curriculum learning method based on input blur. Best viewed in color.}
\label{fig_blur_curriculum}
\vspace*{-0.5cm}
\end{figure}

\noindent
{\bf Curriculum by input blur.}
One strategy to perform curriculum learning without having a way to estimate the difficulty of data samples is to purposely simplify the inputs. In this paper, we propose to artificially simplify the inputs, reducing the amount of information, by blurring the images using a Gaussian filter. We control the amount of blur by adjusting the parameter $\sigma$ with respect to the training stage. In the early training stages, we use a higher value for $\sigma$, blurring the input images in a more aggressive manner. We conjecture that blurred images are easier to align as the blur hides away fine misalignment errors. As the model begins to learn to align blurry images, we gradually reduce $\sigma$ until it becomes $0$ in the later training stages. Hence, at the end of the training process, the neural model is trained on original images. Our curriculum learning strategy is illustrated in Figure~\ref{fig_blur_curriculum}.

\noindent
{\bf Curriculum dropout.} Dropout~\cite{srivastava2014dropout} was introduced as a regularization technique, to prevent neural networks from overfitting. The idea behind dropout is to randomly deactivate certain neurons in a neural network during training, leading to a distributed and more robust representation. In the original formulation, the dropout rate is fixed. Curriculum dropout was introduced in~\cite{morerio2017curriculum} as an adaptive dropout that decreases the rate of dropout in later stages of training up to some minimum value that is set in advance. Since the baseline VTN network does not use dropout, we set the minimum dropout rate to $0.0$.

\noindent
{\bf Curriculum by smoothing.} Similar to our curriculum learning strategy that blurs the examples in the initial stages of training, curriculum by smoothing~\cite{sinha2020curriculum} applies a low-pass filter to convolutional filters. Applying low-pass filters in the early stages of training reduces the noise propagated through the network, improving convergence. As the training progresses, more and more high frequency data is let through the network. In \cite{sinha2020curriculum}, a 2D Gaussian kernel is used as the low-pass filter.
% \begin{equation}\label{eq_2d_gauss}
% k_{G_{\sigma}^{2D}}(x, y, \sigma) = \frac{1}{2\pi\sigma^2}\cdot exp\left(-\frac{x^2 + y^2}{2\sigma^2}\right),
% \end{equation}
Since, we are dealing with 3D scans as input, we need to apply a 3D Gaussian kernel:
\begin{equation}\label{eq_3d_gauss}
k_{G_{\sigma}^{3D}}(x, y, z, \sigma) = \frac{1}{\left(\sqrt{2\pi}\sigma\right)^3}\cdot exp\left(-\frac{x^2 + y^2 + z^2}{2\sigma^2}\right),
\end{equation}
where $\sigma$ controls the amount of blur.

% We took advantage of the fact that the Gaussian blur filter is a linear filter that can be decomposed as a 2D filter $f$ (for the first two axes $x$ and $y$) and a 1D filter $g$ (for the third axis $z$). Applying these filters successively is equivalent to applying the 3D filter equal to $f\cdot g$. 

A typical 3D convolutional network for image registration, such as the 1-cascade VTN~\cite{zhao2019unsupervised,zhao2019recursive}, is constructed of several convolutional blocks of the following form:
\begin{equation}\label{eq_vtn_blocks}
h_i = \mbox{activation}\left(w_i * h_{i-1}\right),
\end{equation}
where $h_{i-1}$ and $h_i$ are the 4D input and output tensors of block $i$, $w_i$ are the learnable parameters of the convolutional block and $*$ denotes the 3D convolution operation. For VTN, the activation function chosen by Zhao et al.~\cite{zhao2019recursive} is leaky ReLU. The smoothing is applied as follows:
\begin{equation}\label{eq_vtn_blocks}
h_i = \mbox{activation}\left(k_{G_{\sigma}^{3D}} \hat{*} (w_i * h_{i-1})\right),
\end{equation}
where $k_{G_{\sigma}^{3D}}$ is a 3D Gaussian kernel and $\hat{*}$ denotes the separable 3D convolution operation. We emphasize that it is not straightforward to apply a 3D Gaussian filter in commonly used libraries for deep neural network training. Indeed, the implementation requires a separable 3D convolution to apply the 3D kernel $k_{G_{\sigma}^{3D}}$ on a 4D input tensor. Another disadvantage of this method is the added model complexity and the additional training time required for applying Gaussian filters on all convolutional layers.
In this regard, our novel approach based on blurring the 3D input scans represents a less cumbersome solution to achieve the same result. % Nevertheless, we will make our implementation of curriculum by smoothing for 3D CNNs available as open source, along with the other curriculum learning strategies.

% The 1-cascade VTN~\cite{zhao2019unsupervised, zhao2019recursive} is constructed of several convolutional blocks of the following form:
% \begin{equation}\label{eq_vtn_blocks}
% h_i = \mbox{LeakyReLU}\left(w_i * h_{i-1}\right),
% \end{equation}
% where $h_{i-1}$ and $h_i$ are the 4D input and output tensors of block $i$, $w_i$ are the learnable parameters of the convolutional block and $*$ denotes the 3D convolution operation. The smoothing is applied as follows:
% \begin{equation}\label{eq_vtn_blocks}
% h_i = \mbox{LeakyReLU}\left(k_{G_{\sigma}^{3D}} \hat{*} (w_i * h_{i-1})\right),
% \end{equation}
% where $k_{G_{\sigma}^{3D}}$ is a 3D Gaussian kernel and $\hat{*}$ denotes the separable 3D convolution operation. We emphasize that it is not straightforward to apply a 3D Gaussian filter in commonly used libraries for deep neural network training. Indeed, the implementation requires a separable 3D convolution to apply the 3D kernel $k_{G_{\sigma}^{3D}}$ on a 4D input tensor. In this regard, our novel approach based on blurring the 3D input scans represents a less cumbersome solution to achieve the same result. Nevertheless, we will make our implementation of curriculum by smoothing for 3D CNNs available as open source, along with the other curriculum learning strategies.

\vspace{-0.25cm}
\section{Experiments}
\label{sec:experiments}
\vspace{-0.15cm}

\begin{table*}[!t]
%\small{
\caption{Dice and Jaccard scores on the SLIVER data set. Three curriculum learning strategies are compared with the conventional training regime. The best results are highlighted in bold. The times required for executing a step with each training strategy are measured on a Google Colab machine with an Intel(R) Xeon(R) CPU at 2.20GHz with 12GB of RAM and an Nvidia Tesla P100 GPU with 16GB of VRAM.}\label{tab_results_sliver}
\vspace{-0.2cm}
\begin{center}
\begin{tabular}{|l|c|c|c|c|c|}
\hline
Method                                              & \#Steps   &  Time per step (seconds)  & Dice          & Jaccard   \\    
\hline
\hline
1-cascade VTN~\cite{zhao2019recursive}              & $40000$   & $0.187$                   & $0.90268$     & $0.82365$ \\  
\hline 
1-cascade VTN based on curriculum by input blur     & $40000$   & $0.193$               & $\mathbf{0.91037}$& $\mathbf{0.83636}$ \\  
1-cascade VTN based on curriculum dropout           & $40000$   & $0.195$                   & $0.90097$     & $0.82088$ \\  
1-cascade VTN based on curriculum by smoothing      & $40000$   & $0.276$                   & $0.90990$     & $0.83557$ \\  
\hline
\end{tabular}
\end{center}
%}
\vspace{-0.4cm}
\end{table*}

\noindent
{\bf Data sets.} We train and evaluate the models on the same liver data sets as~\cite{zhao2019recursive}. More specifically, the training is performed on the MSD~\cite{msd2018} and BFH~\cite{zhao2019unsupervised} data sets. The validation is conducted on the LiTS~\cite{lits2018} data set, while the testing is performed on the SLIVER~\cite{heimann2009comparison} data set. MSD contains various types of CT scans of liver tumors (70 scans), hepatic vessels (443 scans) and pancreas tumors (420 scans). BHF contains 92 scans. There are no annotations for MSD and BHF (the training is unsupervised). LiTS contains 131 liver scans with ground-truth segmentations. Similarly, SLIVER contains 20 scans with ground-truth liver segmentations.
% Brain data sets contained ADNI~\cite{mueller2005ways}, ABIDE~\cite{di2014autism}, ADHD~\cite{bellec2017neuro} that were used for training and LPBA~\cite{shattuck2008construction} for evaluation. 
We used the pre-processed data sets provided by Zhao et al.~\cite{zhao2019recursive}. The pre-processing includes cropping and resampling into volumes of $128\times128 \times 128$ voxels.

\noindent
{\bf Baseline.} In our experiments, we consider the 1-cascade recursive VTN~\cite{zhao2019unsupervised, zhao2019recursive} as the base model, which we trained for a number of 40,000 iterations on randomly generated mini-batches of four samples each.

\noindent
{\bf Parameter tuning.}
We set the training hyperparameters according to \cite{zhao2019recursive}, regardless of the fact that the VTN is trained with conventional or curriculum learning. However, the curriculum learning strategies have additional hyperparameters that need to be set, as described below. All curriculum strategies are applied over the first 20,000 training iterations, using a linear curriculum scheduler. Then, in the last 20,000 iterations, the conventional training regime is resumed. For curriculum by input blur, we apply a Gaussian kernel starting with a $\sigma$ of $1.0$ that linearly decreases to $0.0$ during the first 20,000 training iterations. Analogously, for curriculum by smoothing, the parameter $\sigma$ of the Gaussian filter is decreased linearly from $1.0$ to $0.0$ in the first 20,000 iterations. For curriculum dropout, we linearly decrease the dropout rate from $0.5$ to $0.0$ in the first 20,000 training iterations.

\begin{figure}[!t]
\begin{center}
\includegraphics[width=1.0\linewidth]{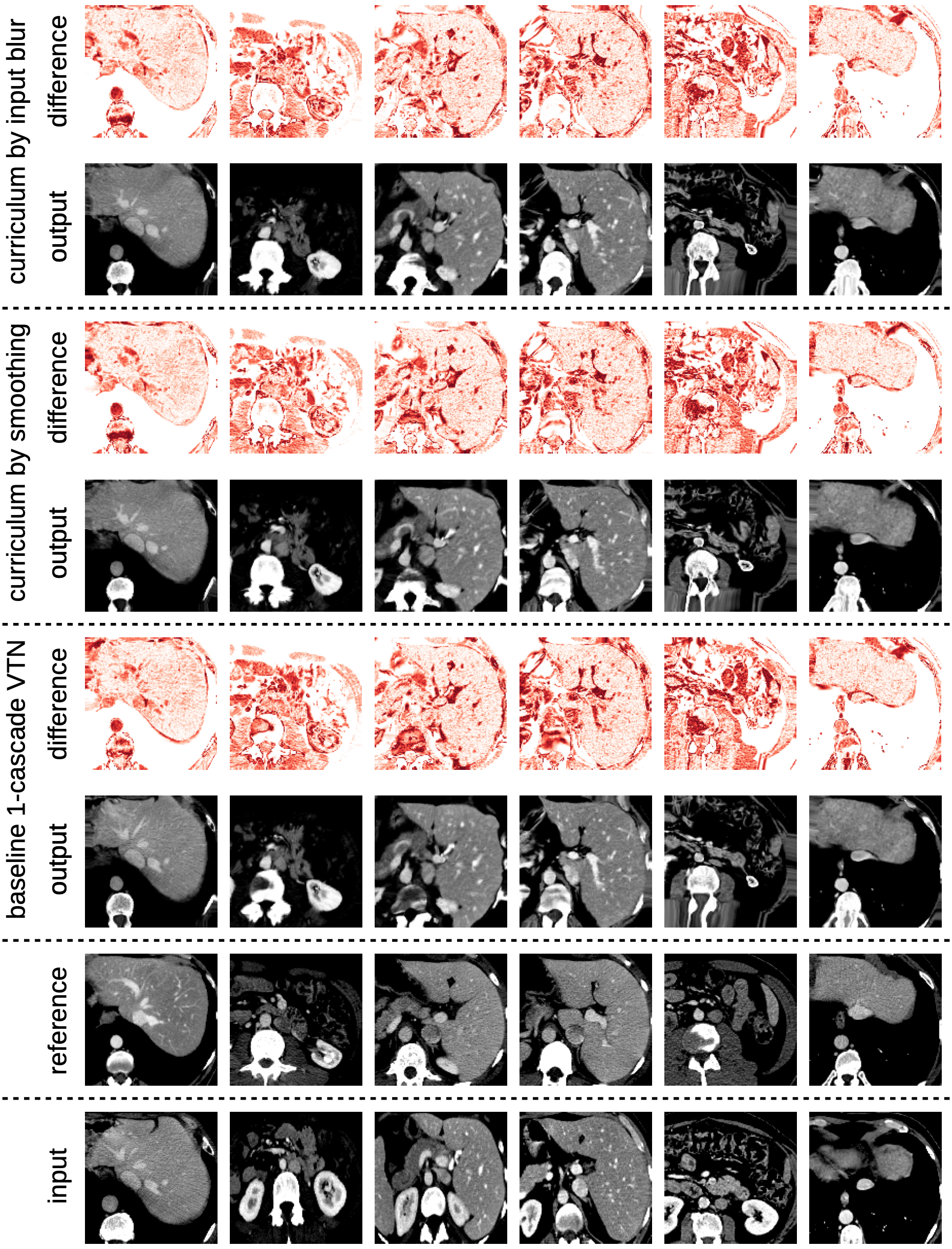}
\end{center}
\vspace{-0.7cm}
\caption{Comparative results of the baseline 1-cascade VTN versus two of our best curriculum learning strategies. The mean absolute differences between the reference and the warped output are represented on a white-to-red scale, where the red intensity is proportional to the magnitude of the difference. Shown samples are 2D excerpts from full 3D scans of different patients. Best viewed in color.}
\label{fig_results}
\vspace*{-0.3cm}
\end{figure}

\noindent
{\bf Evaluation metrics.}
We employ the evaluation framework provided in~\cite{zhao2019recursive} for a direct comparison with the baseline VTN. In the evaluation framework, the considered metrics are the Dice coefficient and the Jaccard index. The Dice coefficient measures the mean overlap as the intersection between the warped and target volumes divided by their mean volume. The intersection-over-union (IoU), also known as the Jaccard index, is an alternative way to measure the overlap as the intersection between the warped and target volumes divided by their union. The two metrics are computed as follows:
\begin{equation}\label{eq_Dice_metric}
\mbox{Dice}(A, B) = 2 \cdot \frac{|A \cap B|}{|A| + |B|}; \mbox{Jaccard}(A, B) = \frac{|A \cap B|}{|A \cup B|}.
\end{equation}
% \begin{equation}\label{eq_Dice_metric}
% \mbox{Jaccard}(A, B) = \frac{|A \cap B|}{|A \cup B|}.
% \end{equation}
For both metrics, higher values indicate better performance.

\noindent
{\bf Results.}
We report the results on the SLIVER data set in Table~\ref{tab_results_sliver}. First, we observe that two of the studied curriculum learning methods achieve superior results compared to the conventional training regime. These are curriculum by input blur and curriculum by smoothing. Their improvements are above $0.72\%$ in terms of the Dice coefficient and above $1.19\%$ in terms of the Jaccard index. We notice that curriculum dropout is not useful in our case. In a set of extra experiments, we observed that dropout alone also degrades performance. This might explain why curriculum dropout is rendered ineffective for the 1-cascade VTN.
In Figure~\ref{fig_results}, we show a series of examples in which the output of the baseline 1-cascade VTN is compared with the outputs of the 1-cascade VTNs based on curriculum by smoothing and curriculum by input blur, respectively. Consistent with the results presented in Table~\ref{tab_results_sliver}, we observe that curriculum learning helps the network to minimize the differences between the warped outputs and the reference images depicted in Figure~\ref{fig_results}.

While curriculum by input blur is the top scoring approach, we emphasize that the training time compared to curriculum by smoothing is significantly lower, being very close to the training time of the baseline VTN. This happens because blurring the input images is very fast. Curriculum by smoothing applies a 3D Gaussian blur on all convolutional filters, which is more time consuming. Curriculum dropout is as fast as our approach based on curriculum by input blur, but, as noted earlier, curriculum dropout does not bring any performance improvements over the baseline VTN. In conclusion, curriculum by input blur provides the best trade-off between accuracy and speed.

% \begin{table*}[!t]
% \small{
% \caption{Dice and Jacc scores on LBPA validation data set}\label{tab_results_lbpa}
% \vspace{0cm}
% \begin{center}
% \begin{tabular}{|l|c|c|c|c|}
% \hline
% Method                                                          & Steps         & Dice                      & Jacc   \\    
  
% \hline
% 1-cascade VTN~\cite{zhao2019recursive}                          & $40000$       & $0.6828070878982544$      & $0.5261470675468445$ \\  
% \hline 
% Blurred input curriculum 1-cascade VTN                          & $40000$       & $0.6831824403762817$      & $0.52890940284729$ \\  
% Curriculum by smoothing 1-cascade VTN                           & $40000$       & $0.6830806732177734$      & $0.52647864818573$ \\  
% \hline
% \end{tabular}
% \end{center}
% }
% \vspace{-0.8cm}
% \end{table*}

%We report the results on the brain data in the table ~\ref{tab_results_lbpa}. Convolutional filter deblurring curriculum achieved better results than baseline after $40000$ steps with a Dice score of $0.6830806732177734$ compared to the baseline Dice score of $0.6828070878982544$.

\vspace{-0.25cm}
\section{Conclusion}
\label{sec:conclusion}
\vspace{-0.15cm}

In this work, we studied the possibility of employing a series of curriculum learning methods for medical image registration. Additionally, we proposed a novel and effective curriculum learning regime based on input blur. Our empirical results showed that curriculum learning can bring performance gains to a state-of-the-art unsupervised medical image registration network \cite{zhao2019recursive}. In future work, we aim to apply curriculum learning strategies on additional medical image registration frameworks and evaluate the resulting models on multiple benchmarks.

% Curriculum by de-blurring~\cite{sinha2020curriculum} and curriculum by increasing the complexity of training examples have proven to achieve increased training performance, reducing the training time on both data set types. This work shows that curriculum methods can be successfully applied to medical image registration tasks.

\noindent
{\bf Acknowledgments.} The research leading to these results has received funding from the NO Grants 2014-2021, under project ELO-Hyp contract no. 24/2020.

\vspace{-0.15cm}

% \vfill\pagebreak

\bibliographystyle{IEEEbib}
\bibliography{references}

\end{document}